\documentclass{article} 
\usepackage{nips12submit_e,times}

\usepackage{graphicx} 
\usepackage{subfigure} 

\usepackage{amsmath,amssymb} 
\usepackage{epsfig}
\usepackage{bm}
\usepackage{color}
\usepackage{url}
\usepackage{enumitem}

\usepackage[]{hyperref}

\title{Efficient Learning of Domain-invariant Image Representations}
\author{
Judy Hoffman\\
UCB EECS \& ICSI\\
\texttt{jhoffman@eecs.berkeley.edu} \\
\AND
Erik Rodner \\
UCB EECS \& ICSI\\
\texttt{erik.rodner@gmail.com} \\
\And
Jeff Donahue \\
UCB EECS \& ICSI\\
\texttt{jdonahue@eecs.berkeley.edu} \\
\And
Trevor Darrell \\
UCB EECS \& ICSI\\
\texttt{trevor@eecs.berkeley.edu}
\And
Kate Saenko \\
University of Massachusetts, Lowell\\
\texttt{saenko@cs.uml.edu} \\
}

\nipsfinalcopy
\begin{document} 

\maketitle

\newcommand{\fix}{\marginpar{FIX}}
\newcommand{\new}{\marginpar{NEW}}

\newcommand{\eqn}[1]{\begin{eqnarray*} #1 \end{eqnarray*}}
\newcommand{\thm}[2]{\begin{description}  \item[Theorem #1:]  #2  \end{description}}
\newcommand{\cor}[2]{\begin{description}  \item[Corollary #1:]  #2  \end{description}}
\newcommand{\todo}[1]{\textcolor{red}{TODO: #1}}
\newcommand{\todoN}[2]{\textcolor{green}{\bf #1}\textcolor{red}{ - #2}}

\begin{abstract} 
We present an algorithm that learns representations which explicitly compensate for domain mismatch and which can be efficiently realized as linear classifiers. Specifically, we form a linear transformation that maps features from the target (test) domain to the source (training) domain as part of training the classifier. We optimize both the transformation and classifier parameters jointly, and introduce an efficient cost function based on misclassification loss.
Our method combines several features previously unavailable in a single algorithm: multi-class adaptation through representation learning, ability to map across heterogeneous feature spaces, and scalability to large datasets. We present experiments on several image datasets that demonstrate improved accuracy and computational advantages compared to previous approaches.
\end{abstract} 

\vspace{-0.3cm}
\section{Introduction}
\vspace{-0.3cm}
We address the problem of learning domain-invariant image representations for multi-class classifiers. The ideal image representation often depends not just on the task but also on the domain. Recent studies have demonstrated a significant degradation in the performance of state-of-the-art image classifiers when input feature distributions change due to different image sensors and noise conditions \cite{ref:eccv_saenko}, pose changes \cite{ref:farhadi}, a shift from commercial to consumer video \cite{ref:duan,ref:duan10}, and, more generally, training datasets biased by the way in which they were collected~\cite{ref:Efros-dataset-bias-cvpr2011}. Learning adaptive representations for linear classifiers is particularly interesting as they are efficient and prevalent in vision applications, with fast linear SVMs forming the core of some of the most popular object detection methods~\cite{ref:DPM,ref:BourdevMalikICCV09}.

\begin{figure}[t]
\centering
\includegraphics[width=\linewidth]{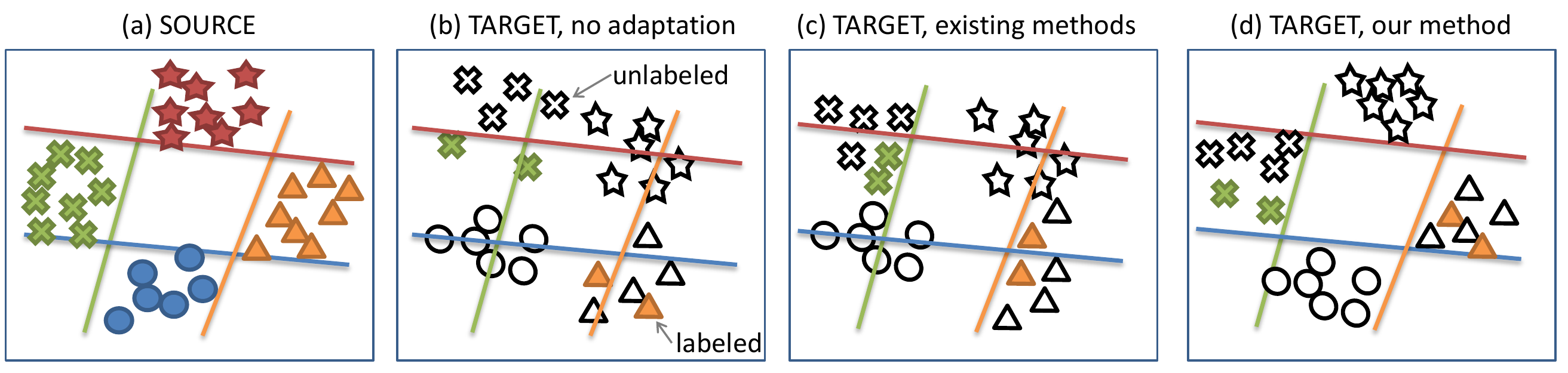} 
\caption{(a) Linear classifiers (shown as decision boundaries) learned for a four-class problem on a fully labeled source domain. (b) Problem: classifiers learned on the source domain do not fit the target domain points shown here due to a change in feature distribution. (c) Existing SVM-based methods only adapt the features of classes with labels (crosses and triangles). (d) Our method adapts all points, including those from classes without labels, by transforming all target features to a new domain-invariant representation.}
\label{fig:concept} 
\end{figure}

Previous work proposed to adapt linear SVMs~\cite{ref:asvm_yang,ref:asvm_li,ref:pmt}, learning a perturbation of the source hyperplane by minimizing the classification error on labeled target examples for each binary task. These perturbations can be thought of as new feature representations that correct for the domain change. The recent HFA method~\cite{ref:duan_icml12} learns both the perturbed classifier and a latent domain-invariant feature representation, allowing domains to have heterogeneous features with different dimensionalities. However, existing SVM-based methods 
are limited to learning a separate representation for each binary problem and cannot transfer a common, class-independent component of the shift (such as global lighting change) to unlabeled categories, as illustrated in Figure~\ref{fig:concept}. Additionally, the HFA algorithm cannot be solved in linear space and therefore scales poorly to large datasets. 

Recently proposed feature adaptation methods~\cite{ref:eccv_saenko,ref:farhadi,ref:kulis_cvpr11,
ref:Chellappa_DAunsupervised_ICCV2011,ref:translated, ref:gong12_gfk} offer a solution by learning a category-independent feature transform that maps target features into the source, pooling all training labels across categories. This enables multi-class adaptation, i.e.~transferring the category-independent component of the domain-invariant representation to unlabeled categories. For example, a map learned on the labeled ``triangle'' class in Figure~\ref{fig:concept} can also be used to map the unlabeled ``star'' class to the source domain. 
An additional advantage of the asymmetric transform method ARC-t~\cite{ref:kulis_cvpr11} over metric learning~\cite{ref:eccv_saenko} or the recently proposed Geodesic Flow Kernel (GFK)~\cite{ref:gong12_gfk}, is that, like HFA~\cite{ref:duan_icml12}, ARC-t can map between heterogeneous feature spaces.
However, ARC-t has two major limitations: First, the feature learning does not optimize the objective function of a strong, discriminative classifier directly; rather, it maximizes some notion of similarity between the transformed target points and points in the source. Second, it does not scale well to domains with large numbers of points due to the high number of constraints, which is proportional to the product of the number of labeled data points in the source and target.

In this paper, we present a novel technique that combines the 
desirable aspects of recent methods in a single algorithm, which we call Max-Margin Domain Transforms, or MMDT for short. MMDT uses an asymmetric (non-square) transform $W$ to map
target features $x$ to a new representation $Wx$ maximally aligned with the source, learning the transform jointly on all categories for which target labels are available (Figure~\ref{fig:concept}(d)).
MMDT provides a way to adapt max-margin classifiers in a multi-class manner, by learning a shared component of the domain shift as captured by the feature transformation $W$. Additionally, MMDT can be optimized quickly in linear space, making it a feasible solution for problem settings with a large amount of training data.

The key idea behind our approach is to simultaneously learn both the projection of the target features into the source domain and the classifier parameters themselves, using the same classification loss to jointly optimize both.

Thus our method learns a feature representation that combines the strengths of max-margin learning with the flexibility of the feature transform.  Because it operates over the input features, it can generalize the learned shift in a way that parameter-based methods cannot. On the other hand, it overcomes the two flaws of the ARC-t method: by optimizing the classification loss directly in the transform learning framework, it can achieve higher accuracy; furthermore, replacing similarity constraints with more efficient hyperplane constraints significantly reduces the training time of the algorithm and learning a transformation directly from target to source allows optimization in linear space. 
\begin{table*}[t]
\begin{center}
\begin{tabular}{|l || c | c  | c | c|}
\hline
                       & ARC-t \cite{ref:kulis_cvpr11} & HFA \cite{ref:duan_icml12} & GFK \cite{ref:gong12_gfk} & MMDT (ours) \\
\hline
multi-class            & \textbf{yes}   & no  & \textbf{yes} & \textbf{yes} \\
large datasets         & no    & no  & \textbf{yes} & \textbf{yes} \\
heterogeneous features & \textbf{yes}   & \textbf{yes} & no  & \textbf{yes} \\
optimize max-margin objective & no & \textbf{yes} & no & \textbf{yes}\\
\hline
\end{tabular}
\end{center}
\caption{Unlike previous methods, our approach is able to simultaneously learn muti-class representations that can transfer to novel classes, scale to large training datasets, and handle different feature dimensionalities.}
\label{table:advantages}
\end{table*}

The main contributions of our paper can be summarized as follows (also see Table~\ref{table:advantages}):
\begin{itemize}
\itemsep4pt
\item Experiments show that MMDT in linear feature space outperforms competing methods in terms of multi-class accuracy even compared to previous kernelized methods.

\item MMDT learns a representation via an asymmetric category independent transform. Therefore, it can adapt features even when the target domain does not have any labeled examples for some categories and when the target and source features are not equivalent.

\item The optimization of MMDT is scalable to large datasets because the number of constraints to optimize is linear in the number of training data points and because it can be optimized in linear feature space.

\item Our final iterative solution can be solved using standard QP packages, making MMDT easy to implement.
\end{itemize}

\section{Related Work}
\vspace{-0.3cm}
Domain adaptation, or covariate shift, is a fundamental problem in machine learning, and has attracted a lot of attention in the machine learning and natural language community, e.g.~\cite{ref:blitzer,ref:daume,ref:bendavid2007,ref:jiang2007a} (see \cite{ref:jiang_survey} for a comprehensive overview). 
It is related to multi-task learning but differs from it in the following way: in domain adaptation problems, the distribution over the features $\mathrm{p}(X)$ varies across domains while the output labels $Y$ remain the same; in multi-task learning or knowledge transfer, $\mathrm{p}(X)$ stays the same (single domain) while the output labels vary (see \cite{ref:jiang_survey} for more details). In this paper, we perform multi-task learning \textit{across domains}, i.e. both $\mathrm{p}(X)$ and the output labels $Y$ can change between domains.
 
Domain adaptation has been gaining considerable attention in the vision community. Several SVM-based approaches have been proposed for image domain adaptation, including: weighted combination of source and target SVMs and  transductive SVMs applied to adaptation in~\cite{BergamoTorresani10}; the feature replication method of~\cite{ref:daume}; Adaptive SVM~\cite{ref:asvm_yang,ref:asvm_li}, where the source model parameters are adapted by adding a perturbation function, and its successor PMT-SVM~\cite{ref:pmt}; Domain Transfer SVM~\cite{ref:duan}, which learns a target decision function while reducing the mismatch in the domain distributions; and a related method~\cite{ref:duan10} based on multiple kernel learning. In the linear case, feature replication~\cite{ref:daume} can be shown to decompose the learned  parameter into $\theta=\hat{\theta} +\theta'$, where $\hat{\theta}$ is shared by all domains~\cite{ref:jiang}, in a similar fashion to adaptive SVMs. 

Several authors considered learning feature representations for unsupervised and transfer learning~\cite{Bengio12deep}, and for domain adaptation~\cite{ref:bendavid2007,Blitzer:2006}.
For visual domain adaptation, transform-based adaptation methods~\cite{ref:eccv_saenko,ref:kulis_cvpr11,ref:Chellappa_DAunsupervised_ICCV2011,ref:farhadi,ref:translated,ref:duan_icml12} have recently been proposed. These methods attempt to learn a perturbation over the feature space rather than a class-specific perturbation over the model parameters, typically in the form of a transformation matrix/kernel. 
The most closely related are the ARC-t method~\cite{ref:kulis_cvpr11}, which learns a transformation that maximizes similarity constraints between points in the source and those projected from the target domain, and the recent HFA method~\cite{ref:duan_icml12}, which learns a transformation both from the source and target into a common latent space, as well as the classifier parameters.
Another related method is the recently proposed GFK~\cite{ref:gong12_gfk}, which computes a symmetric kernel between source and target points based on geodesic flow along a latent manifold. We will present a detailed comparison to these three methods in the next section.

\section{Max-Margin Domain Transforms}
\vspace{-0.3cm}
\label{sec:maxMarginFeatureTransforms}
\newcommand{\mat}[1]{\begin{bmatrix} #1 \end{bmatrix}}

We propose a novel method for multi-task domain adaptation of linear SVMs by learning a target feature representation.  Denote the normal to the affine hyperplane associated with the $k$'th binary SVM as $\theta_k$, $k=1,...,K$, and the offset of that hyperplane from the origin as $b_k$. Intuitively, we would like to learn a new target feature representation that is shared across multiple categories. We propose to do so by estimating a transformation $W$ of the input features, or, equivalently, a transformation $W^T$ of the source hyperplane parameters $\theta_k$.
Let $x_1^s,\dots, x_{n_S}^s$ denote the training points in the source domain ($\mathcal{D}_S$), with labels $y_1^s, \dots, y_{n_S}^s$. Let $x_1^t,\dots, x_{n_T}^t$ denote the labeled points in the target domain ($\mathcal{D}_T$), with labels $y_1^t, \dots, y_{n_T}^t$. 
Thus our goal is to jointly learn 1) affine hyperplanes that separate the classes in the common domain consisting of the source domain and target points projected to the source and 2) the new feature representation of the target domain determined by the transformation $W$ mapping points from the target domain into the source domain. The transformation should have the property that it projects the target points onto the correct side of each source hyperplane.  

For simplicity of presentation, we first show the optimization problem for a binary problem (dropping $k$) with no slack variables. Our objective is as follows:
\begin{eqnarray}
\label{eq:jointObjective}
\min_{W,\theta,b} && \frac{1}{2}||W||_F^2 + \frac{1}{2}||\theta||_2^2\\
\text{s.t.} && y_i^s \left(\mat{x_i^s \\ 1}^T \mat{\theta\\  b}\right) \geq 1 \quad \quad \forall i \in \mathcal{D}_S\\
			&& y_i^t \left( \mat{x_i^t \\ 1}^T W^T \mat{\theta \\ b }\right) \geq  1 \quad \forall i \in\mathcal{D}_T \label{eq:targetConstraints}
\end{eqnarray}
Note that this can be easily extended to the multi-class case by simply adding a sum over the regularizers on all $\theta_k$ parameters and pooling the constraints for all categories. 
The objective function, written as in Equations \eqref{eq:jointObjective}-\eqref{eq:targetConstraints}, is not a convex problem and so is both hard to optimize and is not guaranteed to have a global solution. Therefore, a standard way to solve this problem is to do alternating minimization on the parameters, in our case $W$ and $(\theta,b)$. We can effectively do this because when each parameter vector is fixed, the resulting optimization problem is convex. 

We begin by re-writing Equations \eqref{eq:jointObjective}-\eqref{eq:targetConstraints} for the more general problem with soft constraints and $K$ categories.
Let us denote the hinge loss as: $\mathcal{L}(y,x,\theta)=\max\{0,1-\delta(y,k)\cdot x^T\theta\}$. We define a cost function

{\footnotesize
\begin{eqnarray}
J(W,\theta_k,b_k) &=&
\frac{1}{2}||W||^2_F +  \sum_{k=1}^K \left[\frac{1}{2}||\theta_k||^2_2 \right. \\
&& + C_S\sum_{i=1}^{n_{S}} \mathcal{L}\left(y_i^s, \mat{x_i^s\\1}, \mat{\theta_k\\ b_k}\right) \nonumber  \left. + C_T\sum_{i=1}^{n_{T}}\mathcal{L}\left(y_i^t,W\cdot \mat{x_i^t \\1},\mat{\theta_k\\b_k}\right)\right] \nonumber
\end{eqnarray}
}
where the constant $C_S$ penalizes the source classification error and $C_T$ penalizes the target adaptation error.
Finally, we define our objective function with soft constraints as follows:
\begin{eqnarray}
\min_{W,\theta_k,b_k} J(W,\theta_k,b_k)
\label{eq:obj_soft}
\end{eqnarray}

To solve the above optimization problem we perform coordinate descent on $W$ and $(\theta,b)$.
\begin{enumerate}[leftmargin=0.5cm]
\item Set iteration $j=0$, $W^{(j)}=0$.
\item Solve the sub-problem $(\theta_k^{(j+1)},b_k^{(j+1)} )= \text{arg}\min_{\theta_k,b_k} J(W^{(j)},\theta_k,b_k) $ by solving:
\begin{equation}
\min_{\theta,b}\; \sum_{k=1}^K \left[ \frac{1}{2}||\theta_k||^2_2 \right. + C_S \sum_{i=1}^{n_S} \mathcal{L}\left(y_i^s, \mat{x_i^s\\1}, \mat{\theta_k; \\b_k}\right)  \nonumber \left. + C_T\sum_{i=1}^{n_T}\mathcal{L}\left(y_i^t,W^{(j)}\cdot\mat{x_i^t\\1},\mat{\theta_k\\b_k}\right) \right] \nonumber
	 \end{equation}
	 Notice, this corresponds to the standard SVM objective function, except that the target points are first projected into the source using $W^{(j)}$. Therefore, we can solve this intermediate problem using a standard SVM solver package.

\item Solve the subproblem $W^{(j+1)} = \text{arg}\min_W J(W,\theta^{(j+1)},b^{(j+1)})$ by solving
\begin{eqnarray}
\min_{W} && \frac{1}{2}||W||^2_F + C_T\sum_{k=1}^K\sum_{i=1}^{n_T}\mathcal{L}\left(y_i^t,W\cdot\mat{x_i^t\\1},\mat{\theta_k^{(j+1)}\\b_k^{(j+1)}}\right) \nonumber
\end{eqnarray}
and increment $j$. This optimization sub-problem is convex and is in a form that a standard QP optimization package can solve.
\item Iterate steps 2 \& 3 until convergence. 
\end{enumerate}

It is straightforward to show that both stages (2) and (3) cannot increase the global cost function $J(W,\theta,b)$. Therefore, this algorithm is guaranteed to converge to a local optimum. A proof is included in the supplemental material.

It is important to note that since both steps of our iterative algorithm can be solved using standard QP solvers, the algorithm can be easily implemented. Additionally, since the constraints in our algorithm grow linearly with the number of training points and it can be solved in linear feature space, the optimization can be solved efficiently even as the number of training points grows.

\noindent \textbf{Relation to existing work:} We now analyze the proposed algorithm in the context of the previous feature transform methods ARC-t~\cite{ref:kulis_cvpr11}, HFA~\cite{ref:duan_icml12} and GFK~\cite{ref:gong12_gfk}. ARC-t introduced similarity-based constraints to learn a mapping similar to that in step 3 in our algorithm. This approach creates a constraint for each labeled point $x_i^s$ in the source and labeled point $x_i^t$ in the target, and then learns a transformation $W$ that satisfies constraints of the form $(x_i^s)^T W x_i^t > u$ if the labels of $x_i^s$ and $x_i^t$ are the same, and $(x_i^s)^T W x_i^t < l $ if the labels are different, for some constants $u,l$. 

The ARC-t formulation has two distinct limitations that our method overcomes. First, it must solve $n_S \cdot n_T$ constraints, whereas our formulation only needs to solve $K \cdot n_T$ constraints, for a $K$ category problem. In general, our method scales to much larger source domains than with ARC-t. The second benefit of our max-margin transformation learning approach is that the transformation learned using the max-margin constraints is learned jointly with the classifier, and explicitly seeks to optimize the final SVM classifier objective. While ARC-t's similarity-based constraints seek to map points of the same category arbitrarily close to one another, followed by a separate classifier learning step, we seek simply to project the target points onto the correct side of the learned hyperplane, leading to better classification performance. 

The HFA formulation also takes advantage of the max-margin framework to directly optimize the classification objective while learning transformations. HFA learns the classifier and transformations to a common latent feature representation between the source and target. However, HFA is formulated to solve a binary problem so a new feature transformation must be learned for each category. Therefore,  unlike MMDT, HFA cannot learn a representation that generalizes to novel target categories. 
Additionally, due to the difficulty of defining the dimension of the latent feature representation directly, the authors optimize with respect to a larger combined transformation matrix and a relaxed constraint.
This transformation matrix becomes too large when the feature dimensions in source and target are large so the HFA must usually be solved in kernel space.  This can make the method slow and cause it to scale poorly with the number of training examples. In contrast, our method can be efficiently solved in linear feature space which makes it fast and potentially more scalable. 

Finally, GFK~\cite{ref:gong12_gfk} formulates a kernelized representation of the data that is equivalent to computing the dot product in infinitely many subspaces along the geodesic flow between the source and target domain subspaces. The kernel is defined by the authors to be symmetric and so can not handle source and target domains of different initial dimension. Additionally, GFK does not directly optimize a classification objective. In contrast, our method, MMDT, can handle source and target domains of different feature dimensions via an asymmetric $W$, as well as directly optimizes the classification objective. 

\vspace{-.4cm}
\section{Experiments on Image Datasets}
\vspace{-.3cm}
We now present experiments using the \textit{Office}~\cite{ref:eccv_saenko}, \textit{Caltech256}~\cite{caltech256} and \textit{Bing} \cite{BergamoTorresani10} datasets to evaluate our algorithm according to the following four criteria. 
1) Using a subset of the \textit{Office} and \textit{Caltech256} datasets we evaluate multi-class accuracy performance in a standard supervised domain adaptation setting, where all categories have a small number of labeled examples in the target.
2) Using the full \textit{Office} dataset we evaluate multi-class accuracy for the supervised domain adaptation setting where the source and target have different feature dimensions.
3) Using the full \textit{Office} dataset we evaluate multi-class accuracy in the multi-task domain adaptation setting with novel target categories at test time.
4) Using the \textit{Bing} dataset we assess the ability to scale to larger datasets by analyzing timing performance.
\label{sec:experiments-office}


\paragraph{\textit{Office} Dataset}
The \textit{Office} dataset is a collection of images that provides three distinct domains: \texttt{amazon}, \texttt{webcam}, and \texttt{dslr}. The dataset has 31 categories consisting of common office objects such as chairs, backpacks and keyboards. The \texttt{amazon} domain contains product images (from amazon.com) containing a single object, centered, and usually on a white background. The \texttt{webcam} and \texttt{dslr} domains contain images taken in``the wild" using a webcam or a dslr camera, respectively. They are taken in an office setting and so have different lighting variation and background changes (see Figure~\ref{fig:concept} for some examples.) 
We use the SURF-BoW image features provided by the authors~\cite{ref:eccv_saenko}. More details on how these features were computed can be found in \cite{ref:eccv_saenko}. The available features are vector quantized to 800 dimensions for all domains and additionally for the \texttt{dslr} domain there are 600 dimensional features available (we denote this as \texttt{dslr-600}).

\vspace{-.3cm}
\paragraph{\textit{Office} + \textit{Caltech256} Dataset}
This dataset consists of the 10 common categories shared by the \textit{Office} and \textit{Caltech256} datasets. To better compare to previously reported performance, we use the features provided by \cite{ref:gong12_gfk}, which are also SURF-BoW 800 dimensional features.

\vspace{-.3cm}
\paragraph{\textit{Bing} Dataset}
To demonstrate the effect that constraint set size has on run-time performance, we use the \textit{Bing} dataset from \cite{BergamoTorresani10}, which has a larger number of images in each domain than \textit{Office}. The source domain has images from the Bing search engine and the target domain is from the \textit{Caltech256} benchmark. We run experiments using the first 20 categories and set the number of source examples per category to be 50. We use the train/test split from \cite{BergamoTorresani10} and then vary the number of labeled target examples available from 5 to 25. 

\vspace{-.3cm}
\paragraph{Baselines}
We use the following baselines as a comparison in the experiments where applicable.\footnote{We used the LIBSVM package \cite{CC01a} for kernelized methods and Liblinear \cite{ref:liblinear} package for linear methods.}  
\begin{itemize}[leftmargin=1cm]
\itemsep2pt
\item \textbf{svm$_{s}$}: A support vector machine using source training data.
\item \textbf{svm$_{t}$}: A support vector machine using target training data.
\item \textbf{arc-t}: A category general feature transform method proposed by \cite{ref:kulis_cvpr11}. We implement the transform learning and then apply both a KNN classifier (as originally proposed) and an SVM classifier.
\item \textbf{hfa}: A max-margin transform approach that learns a latent common space between source and target as well as a classifier that can be applied to points in that common space~\cite{ref:duan_icml12}.
\item \textbf{gfk}: The geodesic flow kernel~\cite{ref:gong12_gfk} applied to all source and target data (including test data). Following \cite{ref:gong12_gfk}, we use a 1-nearest neighbor classifier with the kernel.
\end{itemize}

%
\vspace{-0.3cm}
\paragraph{Standard Domain Adaptation Experiment}
\begin{table}
\begin{center}
\begin{tabular}{| l || c | c | c | c | c | c|}

\hline
 &{\bf svm$_s$} & {\bf svm$_t$} & {\bf arct} \cite{ref:kulis_cvpr11}& {\bf hfa} \cite{ref:duan_icml12}& {\bf gfk}\cite{ref:gong12_gfk} & {\bf mmdt} (ours)\\ 
\hline

\hline
a $\rightarrow$ w &   33.9 $\pm$    0.7 &   62.4 $\pm$    0.9 &   55.7 $\pm$    0.9 &   61.8 $\pm$    1.1 &   58.6 $\pm$    1.0 & \textcolor{red} {   64.6 $\pm$    1.2} \\
\hline
a $\rightarrow$ d &   35.0 $\pm$    0.8 & \textcolor{blue}{   55.9 $\pm$    0.8} &   50.2 $\pm$    0.7 &   52.7 $\pm$    0.9 &   50.7 $\pm$    0.8 & \textcolor{red} {   56.7 $\pm$    1.3} \\
\hline
w $\rightarrow$ a &   35.7 $\pm$    0.4 &   45.6 $\pm$    0.7 &   43.4 $\pm$    0.5 &   45.9 $\pm$    0.7 &   44.1 $\pm$    0.4 & \textcolor{red} {   47.7 $\pm$    0.9} \\
\hline
w $\rightarrow$ d &   66.6 $\pm$    0.7 &   55.1 $\pm$    0.8 & \textcolor{red}{   71.3 $\pm$    0.8} &   51.7 $\pm$    1.0 & \textcolor{blue}{   70.5 $\pm$    0.7} &   67.0 $\pm$    1.1 \\
\hline
d $\rightarrow$ a &   34.0 $\pm$    0.3 & \textcolor{blue}{   45.7 $\pm$    0.9} &   42.5 $\pm$    0.5 & \textcolor{blue}{   45.8 $\pm$    0.9} & \textcolor{blue}{   45.7 $\pm$    0.6} & \textcolor{red} {   46.9 $\pm$    1.0} \\
\hline
d $\rightarrow$ w &   74.3 $\pm$    0.5 &   62.1 $\pm$    0.8 & \textcolor{red}{   78.3 $\pm$    0.5} &   62.1 $\pm$    0.7 &   76.5 $\pm$    0.5 &   74.1 $\pm$    0.8 \\
\hline
a $\rightarrow$ c &   35.1 $\pm$    0.3 &   32.0 $\pm$    0.8 & \textcolor{red}{   37.0 $\pm$    0.4} &   31.1 $\pm$    0.6 & \textcolor{blue}{   36.0 $\pm$    0.5} & \textcolor{blue} {   36.4 $\pm$    0.8} \\
\hline
w $\rightarrow$ c & \textcolor{blue}{   31.3 $\pm$    0.4} &   30.4 $\pm$    0.7 & \textcolor{red}{   31.9 $\pm$    0.5} &   29.4 $\pm$    0.6 & \textcolor{blue}{   31.1 $\pm$    0.6} & \textcolor{red} {   32.2 $\pm$    0.8} \\
\hline
d $\rightarrow$ c &   31.4 $\pm$    0.3 &   31.7 $\pm$    0.6 & \textcolor{blue}{   33.5 $\pm$    0.4} &   31.0 $\pm$    0.5 & \textcolor{blue}{   32.9 $\pm$    0.5} & \textcolor{red} {   34.1 $\pm$    0.8} \\
\hline
c $\rightarrow$ a &   35.9 $\pm$    0.4 &   45.3 $\pm$    0.9 &   44.1 $\pm$    0.6 &   45.5 $\pm$    0.9 &   44.7 $\pm$    0.8 & \textcolor{red} {   49.4 $\pm$    0.8} \\
\hline
c $\rightarrow$ w &   30.8 $\pm$    1.1 &   60.3 $\pm$    1.0 &   55.9 $\pm$    1.0 &   60.5 $\pm$    0.9 & \textcolor{red}{   63.7 $\pm$    0.8} & \textcolor{red} {   63.8 $\pm$    1.1} \\
\hline
c $\rightarrow$ d &   35.6 $\pm$    0.7 &   55.8 $\pm$    0.9 &   50.6 $\pm$    0.8 &   51.9 $\pm$    1.1 & \textcolor{red}{   57.7 $\pm$    1.1} & \textcolor{blue} {   56.5 $\pm$    0.9} \\
\hline \hline
mean &   40.0 $\pm$    0.6 &   48.5 $\pm$    0.8 &   49.5 $\pm$    0.6 &   47.4 $\pm$    0.8 & \textcolor{blue}{   51.0 $\pm$    0.7} & \textcolor{red} {   52.5 $\pm$    1.0} \\
\hline
\end{tabular}
\end{center}
\caption{Multi-class accuracy for the standard supervised domain adaptation setting: All results are from our implementation. When averaged across all domain shifts the reported average value for {\bf gfk} was 51.65 while our implementation had an average of 51.0 $\pm$ 0.7. Therefore, the result difference is within the standard deviation over data splits. Red indicates the best result for each domain split. Blue indicates the group of results that are close to the best performing result. The domain names are shortened for space: a: \texttt{amazon}, w: \texttt{webcam}, d: \texttt{dslr}, c: \textit{Caltech256}}
\label{table:samecat_office}
\end{table}

For our first experiment, we use the \textit{Office}+\textit{Caltech256} domain adaptation benchmark dataset to evaluate multi-class accuracy in the standard domain adaptation setting  where a few labeled examples are available for all categories in the target domain. We follow the setup of~\cite{ref:eccv_saenko} and \cite{ref:gong12_gfk}: 20 training examples for \texttt{amazon} source (8 for all other domains as source) and 3 labeled examples per category for the target domain. We created 20 random train/test splits and averaged results across them. 

The multi-class accuracy for each domain pair is shown in Table~\ref{table:samecat_office}. Our method produced the highest multi-class accuracy for 9 out of 12 of the domain shifts and competitively on the other 3 shifts.  This experiment demonstrates that our method achieves a high recognition performance and is able to outperform the most recent domain adaptation algorithms. Our method especially stands out in the settings where the domains are initially very different. The most similar domains in this dataset are \texttt{webcam} and \texttt{dslr} and we see that our algorithm does not perform as well on those two shifts as {\bf gfk}. This fits with our intuition since {\bf gfk} is a 1-nearest neighbor approach and so is more suitable when the domains are initially similar. 

Additionally, an important observation is that our linear method on average outperforms all the baselines, even though they each learn a non-linear transformation.

\vspace{-0.3cm}
\paragraph{Asymmetric Transform Experiment}
Next, we analyze the effectiveness of our asymmetric transform learning by experimenting with the source and target having different feature dimensions. We use the same experimental setup as previously, but use the \textit{Office} dataset and the alternate representation for the \texttt{dslr} domain that is 600-dimensional (denoted as \texttt{dslr-600}). We compare against {\bf svm$_t$}, {\bf arc-t} and {\bf hfa}, the baselines that can handle this scenario. The results are shown in Table~\ref{table:diff_dims}. Again, we find that our method can effectively learn a feature representation for the target domain that optimizes the final classification objective.

\begin{table*}

\begin{center}
\begin{tabular}{|c | c || c | c  | c | c|}
\hline
source & target &{\bf svm$_t$} & \bf{arc-t} & \bf{hfa} & \bf{mmdt} \\
\hline \hline
\texttt{amazon} & \texttt{dslr-600} &52.9 $\pm$ 0.7 & 58.2 $\pm$ 0.6 & 57.8 $\pm$ 0.6 & \textcolor{red}{62.3 $\pm$ 0.8} \\
\hline
\texttt{webcam} & \texttt{dslr-600} &51.8 $\pm$ 0.6 &58.2 $\pm$ 0.7 & 60.0 $\pm$ 0.6 & \textcolor{red}{63.3 $\pm$ 0.5}\\
\hline
\end{tabular}

\end{center}
\caption{Multiclass accuracy results on the standard supervised domain adaptation task with different feature dimensions in the source and target. The target domain is \texttt{dslr} for both cases.}

\label{table:diff_dims}
\end{table*}

\vspace{-0.3cm}
\paragraph{Generalizing to Novel Categories Experiment}

\begin{table}
\begin{center}

\begin{tabular}{| c || c | c  | c | c | c| }
\hline
source & {\bf svm}$_s$ &  {\bf arc-t} & {\bf gfk} & {\bf mmdt} \\
\hline
\hline
\texttt{amazon} & 10.3 $\pm$ 0.6 & 41.4 $\pm$ 0.3 & 38.9 $\pm$ 0.4 & \textcolor{red}{44.6 $\pm$ 0.3}\\
\hline
\texttt{webcam} &51.6 $\pm$ 0.5 & 59.4 $\pm$ 0.4 & \textcolor{red}{62.9 $\pm$ 0.5} & 58.3 $\pm$ 0.5 \\
  \hline
\end{tabular}
\caption{Multiclass accuracy results on the \textit{Office} dataset for the domain shift of \texttt{webcam}$\rightarrow$\texttt{dslr} for target test categories not seen in at training time. Following the experimental setup of \cite{ref:kulis_cvpr11}.  We compare against pmt-svm \cite{ref:pmt} and ARC-t \cite{ref:kulis_cvpr11} using both knn and svm classification.}
\label{table:office-diffcat}
\end{center}
\vspace{-.5cm}
\end{table}

We next consider the setting of practical importance where labeled target examples are not available for all objects. Recall that this is a setting that many category specific adaptation methods cannot generalize to, including {\bf hfa}~\cite{ref:duan_icml12}. Therefore, we compare our results for this setting to the {\bf arc-t}~\cite{ref:kulis_cvpr11} method which learns a category independent feature transform and the {\bf gfk}~\cite{ref:gong12_gfk} method which learns a category independent kernel to compare the domains. 
Following the experimental setup of \cite{ref:kulis_cvpr11}, we use the full \textit{Office} dataset and allow 20 labeled examples per category in the source for \texttt{amazon} and 10 labeled examples for the first 15 object categories in the target (\texttt{dslr}). For the \texttt{webcam}$\rightarrow$\texttt{dslr} shift, we use 8 labeled examples per category in the source for \texttt{webcam} and 4 labeled examples for the first 15 object categories in the target \texttt{dslr}.

The experimental results for the domain shift of \texttt{webcam}$\rightarrow$\texttt{dslr} are evaluated and shown in Table~\ref{table:office-diffcat}; MMDT outperforms the baselines for the \texttt{amazon}$\rightarrow$\texttt{dslr} shift and offers adaptive benefit over {\bf svm$_s$} for the shift from \texttt{webcam} to \texttt{dslr}. As in the first experiment, both {\bf arc-t} and {\bf gfk} use nearest neighbor classifiers on a learned kernel are more suitable to the shift between \texttt{webcam} and \texttt{dslr}, which are initially very similar.

\paragraph{Scaling to Larger Datasets Experiment}

With our last experiment we show that our method not only offers high accuracy performance it also scales well with an increasing dataset size. Specifically, the number of constraints our algorithm optimizes scales linearly with the number of training points. 
Conversely, the number of constraints that need to be optimized for the {\bf arc-t} baseline is quadratic in the number of training points. 

\begin{figure}[t]
\begin{center}
\includegraphics[width=0.49\linewidth]{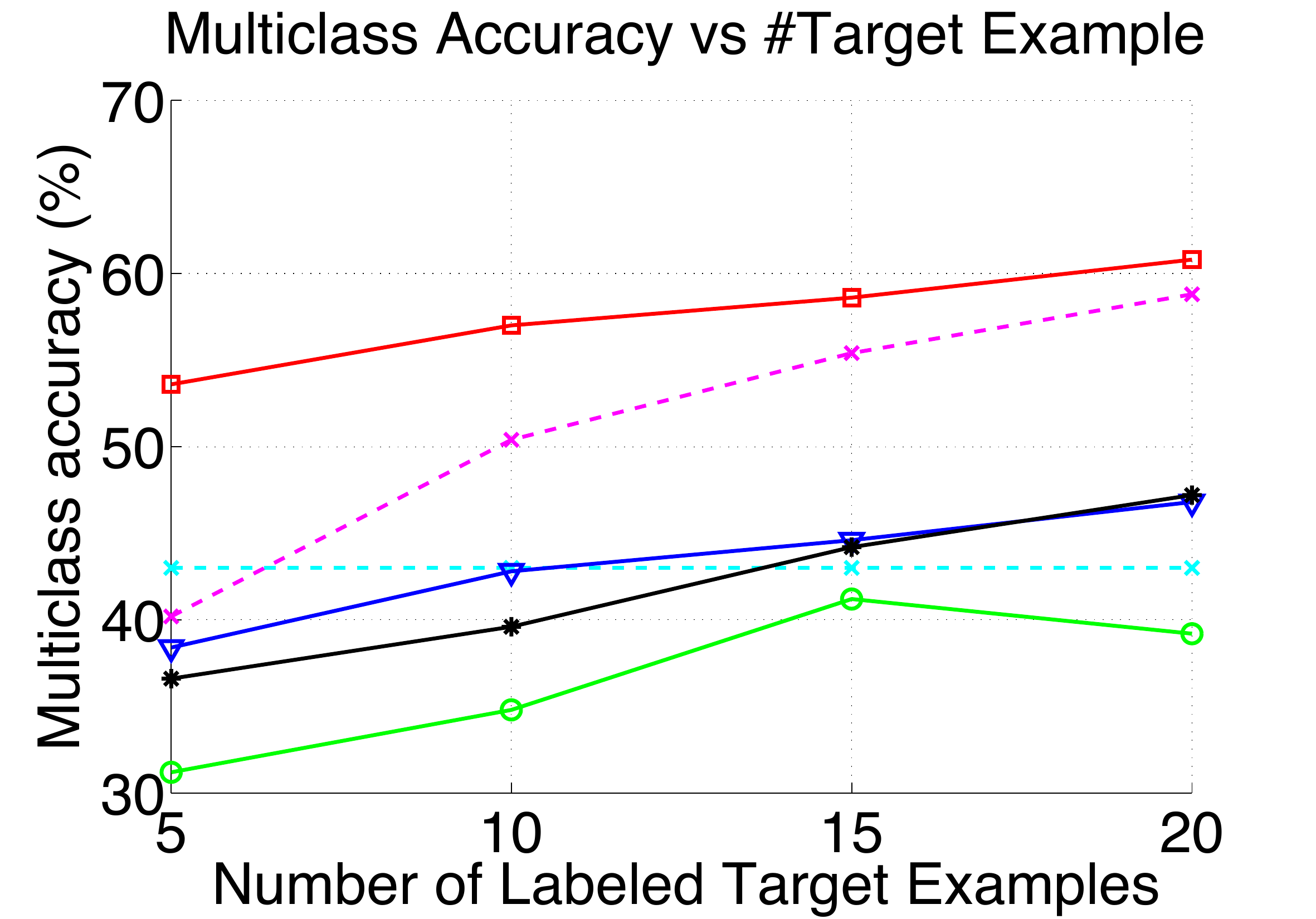}
\includegraphics[width=0.49\linewidth]{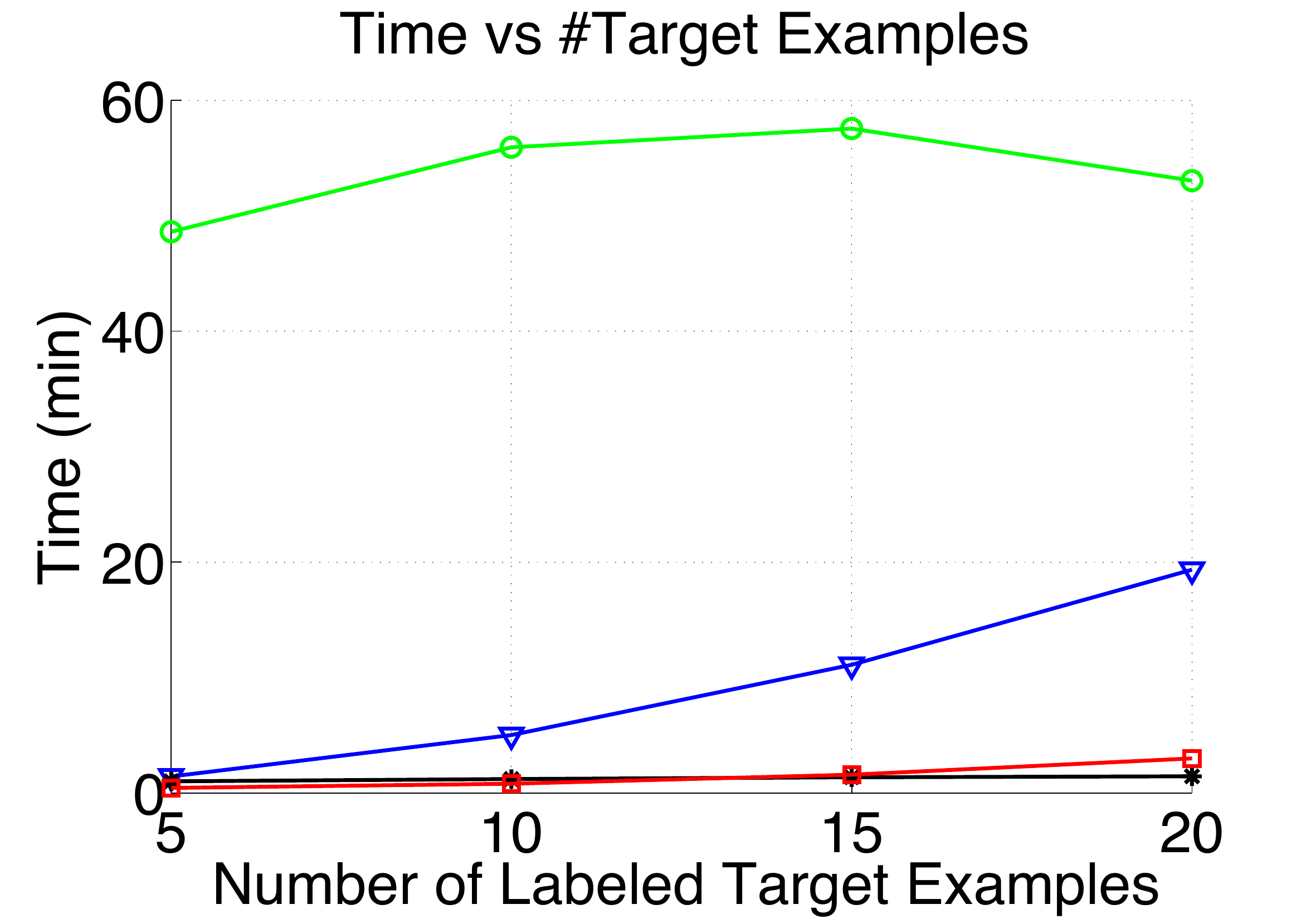}

\includegraphics[width=0.6\linewidth]{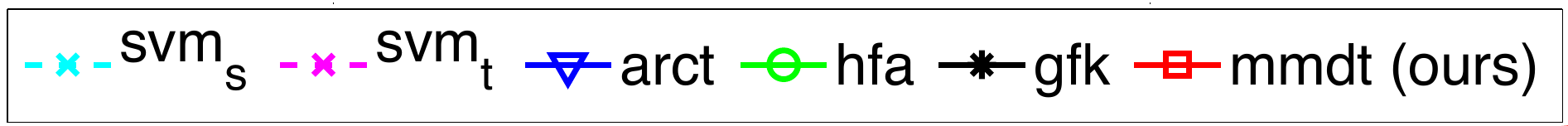}
\end{center}
\vspace{-.5cm}
\caption{Left: multiclass accuracy on the \textit{Bing} dataset using 50 training examples in the source and varying the number of available labeled examples in the target. Right: training time comparison. 
}
\label{fig:bing}
\end{figure}

To demonstrate the effect that constraint set size has on run-time performance, we use the \textit{Bing}~\cite{BergamoTorresani10} dataset, which has a larger number of images in each domain than \textit{Office}. The source domain has images from the Bing search engine and the target domain is from the \textit{Caltech256} benchmark. We run experiments using the first 20 categories and set the number of source examples per category to be 50. We use the train/test split from \cite{BergamoTorresani10} and then vary the number of labeled target examples available from 5 to 20. 
The left-hand plot in Figure \ref{fig:bing} presents multi-class accuracy for this setup. Additionally, the training time of our method (run to convergence) and that of the baselines is shown on the right-hand plot.
 
Our {\bf mmdt} method provides a considerable improvement over all the baselines in terms of multi-class accuracy.  It is also considerably faster than all but the {\bf gfk} method. An important point to note is that both our method and {\bf arc-t} scale approximately linearly with the number of target training points which is empirical verification for our claims. Note that {\bf hfa} and {\bf gfk} do not vary significantly as the number of target training points increases. However, for {\bf hfa} the main bottleneck time is consumed by a distance computation between each pair of training points. Therefore, since there are many more source training points than target, adding a few more target points does not significantly increase the overall time spent for this experiment, but would present a problem as the size of the dataset grew in general.   

\vspace{-0.3cm}
\section{Conclusion}
\vspace{-0.3cm}

In this paper, we presented a feature learning technique for domain adaptation that combines the ability of feature transform-based methods to perform multi-task adaptation  with the performance benefits of directly adapting classifier parameters. 

We validated the computational efficiency and effectiveness of our method using two standard benchmarks used for image domain adaptation. Our experiments show that 1) our method is a competitive domain adaptation algorithm able to outperform previous methods, 2) is successfully able to generalize to novel target categories at test time, and 3) can learn asymmetric transformations. In addition, these benefits are offered through a framework that is scalable to larger datasets and achieves higher classification accuracy than previous approaches.  

So far we have focused on linear transforms because of its speed and scalability; however, our method can also be kernelized to include nonlinear transforms. In future work, we would like to explore the kernelized version of our algorithm and especially experiment with the geodesic flow kernel as input to our algorithm.

\noindent \textbf{Acknowledgements:} This work was supported by NSF grants IIS-1116411 and IIS-1212798, DARPA, and the Toyota Corporation.

\small{
\bibliographystyle{unsrt}
\bibliography{main}
}

\end{document}